\documentclass[10pt,twocolumn,letterpaper]{article}

\usepackage{cvpr}              
\definecolor{cvprblue}{rgb}{0.21,0.49,0.74}
\usepackage[pagebackref,breaklinks,colorlinks,allcolors=cvprblue]{hyperref}

\def\paperID{14814} 
\def\confName{CVPR}
\def\confYear{2026}

\usepackage{graphicx}
\usepackage{multirow}
\usepackage{amsmath}
\usepackage{bm}
\usepackage{bbm}
\usepackage{amssymb}
\usepackage{xspace}

\usepackage{natbib}

\usepackage{caption}   

\title{A Turn Toward Better Alignment: Few-Shot Generative Adaptation with Equivariant Feature Rotation}

\author{Chenghao Xu$^{1}$, Qi Liu$^{2}$, Jiexi Yan$^{3}$, Muli Yang$^{4}$,  Cheng Deng$^{1}$\thanks{Corresponding author} \\
        $^{1}$ Hohai university, China, \\  
        $^{2}$ School of Electronic Engineering, Xidian University, China, \\  
        $^{3}$ School of Computer Science and Technology, Xidian University, China, \\ 
        $^{4}$ Institute for Infocomm Research (I\textsuperscript{2}R), A*STAR, Singapore 
        }

\begin{document}
\maketitle
\begin{abstract}
Few-shot image generation aims to effectively adapt a source generative model to a target domain using very few training images. Most existing approaches introduce consistency constraints—typically through instance-level or distribution-level loss functions—to directly align the distribution patterns of source and target domains within their respective latent spaces. However, these strategies often fall short: overly strict constraints can amplify the negative effects of the domain gap, leading to distorted or uninformative content, while overly relaxed constraints may fail to leverage the source domain effectively.
This limitation primarily stems from the inherent discrepancy in the underlying distribution structures of the source and target domains. The scarcity of target samples further compounds this issue by hindering accurate estimation of the target domain's distribution.
To overcome these limitations, we propose Equivariant Feature Rotation (EFR), a novel adaptation strategy that aligns source and target domains at two complementary levels within a self-rotated proxy feature space. Specifically, we perform adaptive rotations within a parameterized Lie Group to transform both source and target features into an equivariant proxy space, where alignment is conducted. These learnable rotation matrices serve to bridge the domain gap by preserving intra-domain structural information without distortion, while the alignment optimization facilitates effective knowledge transfer from the source to the target domain.
Comprehensive experiments on a variety of commonly used datasets demonstrate that our method significantly enhances the generative performance within the targeted domain. 
\end{abstract}    
\section{Introduction}
\label{sec:intro}

In recent years, there has been an exponential advancement in the field of generative vision tasks, particularly with the advent of deep generative models such as Generative Adversarial Networks (GANs). These models have proven to be remarkably successful in numerous tasks, including but not limited to, natural image synthesis~\cite{brock2018large,karras2019style,zhang2022survey}, image editing~\cite{zhu2017unpaired,cao2021unifacegan}, and image inpainting~\cite{yeh2017semantic,yu2018generative,quan2022image}. The results yielded by GANs have been highly persuasive, demonstrating their capacity for realistic image generation. However, a significant challenge associated with GANs is their requirement for substantial volumes of data during the training process. For instance, popular datasets employed for GAN training, including FFHQ~\citep{karras2019style} and LSUN church~\citep{yu2015lsun}, comprise 70,000 and 126,000 images, respectively. The lack of sufficient training data has been observed to lead to overfitting and collapse of generative models, thereby resulting in suboptimal performance. Consequently, the necessity for large quantities of training data presents a pivotal limitation of GANs, which necessitates prompt attention to enhance the versatility and practicality of these models in real-world applications.

\begin{figure*}[t]
    \centering
\includegraphics[width = 0.98\textwidth]{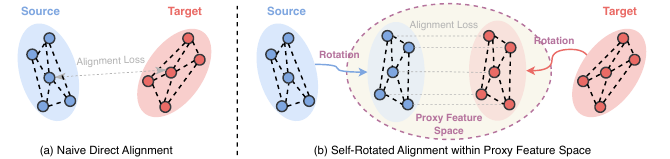}
    \caption{Comparison between the existing naive direct alignment and our self-rotated alignment.}
    \label{fig:intro}
\end{figure*}

In light of this, an increasing number of researchers~\citep{huang2022masked,TGANada,tran2021data,tseng2021regularizing,yang2021data,zhao2020leveraging, duan2024weditgan} are striving to achieve robust image generation in the face of limited training data. A common strategy in this regard involves the adoption of few-shot generative models. This process entails fine-tuning a model that has been pre-trained on a comprehensive dataset from a source domain to accommodate a new domain characterized by limited target data. Through this approach, adaptation methods can produce diverse and realistic images for the target domain, even when faced with as few as ten training images.

The principal challenge in few-shot generative adaptation is to prevent overfitting while maintaining content consistency during the transfer from the source to the target domain. To address this, various loss functions, such as IDC~\citep{CDC} and RSSA~\citep{RSSA}, have been proposed to enhance content preservation and structural coherence throughout the adaptation process. 
Although these approaches have led to notable improvements, directly imposing instance-level and distribution-level alignment between the source and target domains can compromise both training stability and accuracy. This is primarily due to the inherent discrepancy between the underlying distribution patterns of the source and target domains. Furthermore, the limited availability of target samples exacerbates this issue, making it more challenging to accurately represent the target domain’s distribution during adaptation.

In this paper, we introduce a novel perspective based on a proxy feature space and propose the \textbf{Equivariant Feature Rotation} (EFR) method, a new approach designed to enable stable transfer of relevant content information from the source domain while preserving the model’s capacity to acquire style characteristics from the target domain. This is accomplished by aligning the source and target domains at two complementary levels within a self-rotated proxy feature space.
Specifically, we first propose an equivariant self-rotated proxy feature space projection strategy, which performs adaptive rotations within a parameterized Lie Group. 
Instead of directly aligning the source and target data, we transform the source and target domain samples into an equivariant proxy space and perform distribution pattern alignment between source and target domain within this proxy feature space. 
This strategy significantly improves training stability and preserves the structural integrity of the generated distribution, thereby alleviating the negative impact of domain discrepancies.
Following this, we apply both instance-level and distribution-level alignment within the proxy feature space to enforce flexible identity consistency across domains for each instance. Recognizing that direct pairwise alignment may lead to excessive imposition of source domain content, we instead adopt a global distribution pattern alignment based on optimal transport theory. This more relaxed and principled alignment enables effective transfer of semantically meaningful content while avoiding the undesired entanglement of aggressive attribute information from the source domain.

In summary, our contributions are as follows:
\begin{itemize}
	\item  Current losses enforce local pairwise alignment of generated samples in target and source domains, which does not resolve issue of distribution rotation. Therefore, we propose to autonomously rotate target domain's feature distribution to align with source domain within a proxy feature space, thereby ensuring consistency between domains while maintaining local pairwise alignment.
	\item  Additionally, strong cross-domain alignment in existing losses may overemphasize content from source domain, potentially introducing extraneous information and diminishing generative performance. So, we adopt optimal distribution matching strategy based on OT theory.
        \item Extensive experiments on several widely used datasets demonstrate that our method effectively improves the generative quality of the target domain.
\end{itemize}
\section{Related Work}
\label{sec:formatting}

\subsection{Image Generation with Limited Training Data.}
Efforts abound in the realm of instructing a generative model using a limited dataset. A number of previous methodologies have advocated for the application of data augmentation, aimed at curtailing the discriminator's propensity for overfitting. For instance, Zhang and Khoreva~\cite{zhang2019progressive} introduced a progressive augmentation approach. An in-depth exploration of the effects of diverse augmentations during the training process was undertaken by y Zhao et al.~\cite{zhao2020image}, while Tran et al.~\cite{tran2021data} analyzed the theoretical underpinnings of several data augmentations. Moreover, Zhao et al. proposed the utilization of augmentations on both authentic and fabricated images in a differentiable fashion. Karras et al.~\cite{TGANada} conceptualized an adaptive discriminator augmentation, specifically designed to preclude leakage and stabilize the training process. Adding to this, numerous regularizers have been introduced to provide supplementary supervision. For instance, Zhao et al.~\cite{zhao2021improved} incorporated consistency regularization for GANs, exhibiting competitive performance even with limited data. To boost data efficiency, Yang et al.~\cite{yang2021data} integrated contrastive learning into the training framework as an auxiliary task.

\subsection{Few-shot Generative Model Adaptation.}
Transfer learning has emerged as a prevalent technique to enhance a model's generalization capabilities within a specific target domain~\citep{zhuang2020comprehensive,li2023task,pan2009survey}, by capitalizing on knowledge acquired from pre-training on a separate source domain encompassing extensive data. Few-shot image generation~\cite{bounareli2022finding, collins2020editing} entails adapting a pre-trained generative model to a novel target domain characterized by limited data. This established paradigm can be bifurcated into two categories: fine-tuning and regularization methods.

Fine-tuning methods address the issue of an overabundance of trainable parameters by updating only a portion of the model or incorporating additional parameters while maintaining the core model intact. Recent advancements in this domain are AdAM~\citep{AdAM} and RICK~\citep{RICK}, which employ Fisher Information to modulate critical kernels, thereby achieving results on par with regularization-based methodologies. Contrastingly, regularization methods fine-tune all model parameters, imposing penalties on parameter or feature alterations and advocating for the alignment of feature distributions during transfer. For instance, EWC~\citep{EWC} modifies penalties based on Fisher Information to sustain feature consistency between source and target samples. CDC~\citep{CDC} underscores consistency via a loss term. RSSA~\citep{RSSA} incorporates spatial consistency losses to preserve structure. DCL~\citep{DCL} introduces contrastive losses for both generator and discriminator features. Lastly, DWSC~\citep{DWSC} formulates perceptual and contextual losses for varying patch complexities.
\section{Method}

\subsection{Preliminaries}

\paragraph{Problem formulation.} Existing few-shot image generation methods focus on a transfer learning paradigm that leverages a  source generator $\mathcal{G}_s$ pre-trained on a large-scale dataset such as FFHQ~\citep{karras2019style} and then adapts it to a new domain with limited target images. During this adaptation procedure, we fine-tune the source GAN on the limited target data to derive a target generator $\mathcal{G}_t$ as follows:
\begin{equation}
\begin{aligned}
     \mathcal{L}_{G_t} &= -\mathbb{E}_{x \sim \mathcal{T}}[\log( \mathcal{D}_t(\mathcal{G}_t(z)))]  \\
    \mathcal{L}_D  &= \mathbb{E}_{x \sim \mathcal{T}}[\log(1 - \mathcal{D}_t(x))] + \mathbb{E}_{z\sim p(z)}[\log (\mathcal{D}_t(\mathcal{G}_t(z)))],
\end{aligned}
\end{equation}
where $z$ denotes a noise vector sampled from noise distribution $p(z)$ including Gaussian distribution, and $\mathcal{T}$ is the target data distribution. Note that  $\mathcal{D}_t$ is the discriminator corresponding to $\mathcal{G}_t$.
The core goal of few-shot image generation is to capture the inaccessible $\mathcal{T}$.

\begin{figure*}[t]
    \centering
\includegraphics[width = 0.98\textwidth]{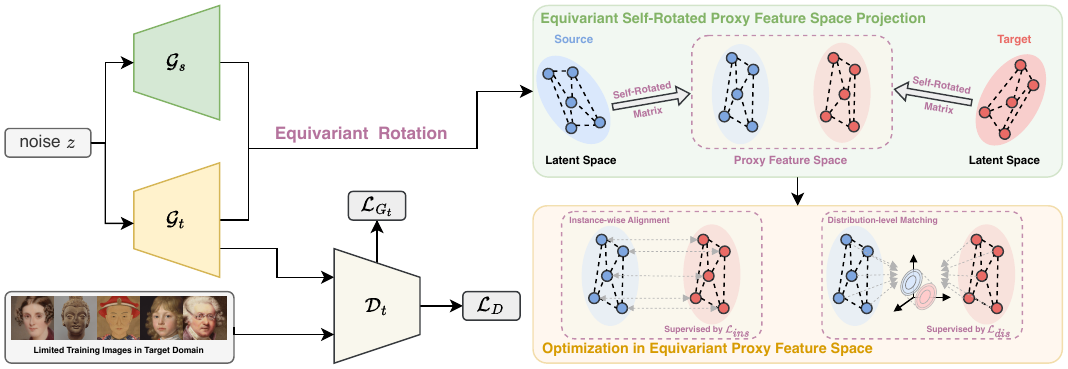}
    \caption{The overall framework of our proposed method.}
    \label{fig:framwork}
\end{figure*}

\paragraph{Main Challenge and Contribution.} To achieve better alignment between the source and target distributional patterns, we propose to impose equivariant feature rotation during optimization.  The overall framework of the proposed EFR is shown in Figure~\ref{fig:framwork}. We first project the source and target data in a common proxy feature space, where the target distributional pattern is self-rotated for better alignment. And then we conduct instance-wise and distribution-level optimization. Specifically, we align the approximate locations of the distribution patterns in source and target domains, improving the training stability. Then, we match the distribution pattern in the target domain with the ones in the source domain from a global perspective, which can effectively maintain identity consistency while preventing undesirable information during generative adaptation.

\subsection{Equivariant Self-Rotated Proxy Feature Space Projection}



Existing few-shot generative adaptation methods primarily focus on preserving the distribution of generated images to match that of the source domain by employing various loss functions~\cite{RSSA, CDC}. This is commonly achieved through a combination of local sample-wise alignment and global distribution-level matching between the source and target domains. 
However, these strategies often suffer from the severe domain gap issue, and hence fail to leverage the source domain effectively,  resulting in distorted or uninformative content.

To address this limitation, we propose an \textit{equivariant self-rotated proxy feature space projection} strategy based on a parameterized Lie group. This approach autonomously rotates the target domain's feature distribution within a proxy feature space to align with that of the source domain, thereby enhancing consistency between the two domains. Specifically, we enforce an equivariant rotation parameterized by an orthogonal matrix $\bm{R}^{*}$ in the proxy feature space, which facilitates better preservation of structural consistency in the distributional patterns across domains.
To better perform orthogonal rotation, we exploit a rotation matrix that lies in a particular orthogonal group, \textit{i.e.}, Lie Group $SO(d)$, which is defined as follows:
\begin{equation}
    SO(d) = \lbrace \bm{R} \in \mathbb{R}^{d\times d} \, | \, \bm{R}^{\top}\bm{R}=\bm{I}, \det \bm{A} =1 \rbrace.
\end{equation}
Note that the standard SGD can not assure that $\bm{R}^{*}$ always be in $SO(d)$ during training. We tend to address this issue in an algebra way~\citep{lezcano2019cheap}, where the Lie Algebra $\mathfrak{so}(d)$ is formed by skew-matrices:
\begin{equation}
    \mathfrak{so}(d) = \lbrace \bm{R} \in \mathbb{R}^{d\times d} \, | \, \bm{R} + \bm{R}^{\top} = 0 \rbrace.
\end{equation}
There exists a mapping $\exp(\cdot): \mathfrak{so}(d)  \rightarrow SO(d)$ defined as:
\begin{equation}
    \exp(\bm{R}) = \bm{I} + \bm{R} + \frac{\bm{R}^2}{2} + \cdots 
\end{equation}
Therefore, the optimization in $SO(d)$ could be transformed into optimization in $\mathfrak{so}(d)$. Furthermore, the Lie Algebra $\mathfrak{so}(d)$ is isomorphic to a linear space. The isomorphism mapping is given by $ \bm{R} \rightarrow \bm{R} - \bm{R}^{\top} $. Consequently, the optimization with orthogonal constraint is transformed into the optimization in $\mathbf{R}^{d\times d}$ as follows:
\begin{equation}
    \min_{\bm{R} \in SO(d)} \mathcal{L}(\bm{R}) \Longleftrightarrow  \min_{\hat{\bm{R}}\in \mathbb{R}^{d\times d}} \mathcal{L}(\exp(\hat{\bm{R}}-\hat{\bm{R}}^{\top}))
    \label{eq:trans}
\end{equation}
In the above formulation, the optimization with orthogonal constraint is transformed into the optimization in $\mathbb{R}^{d\times d}$. Therefore,  we can straightly adopt standard optimization techniques such as SGD~\citep{SGD} and Adam~\citep{ADAMopt} for the right side of Eq.(\ref{eq:trans}). 

During optimization, simultaneously rotating both the source and target domains is equivalent to rotating only a single domain. Furthermore, rotating a single domain is computationally more efficient. Therefore, in the subsequent optimization process, we apply rotation exclusively to the target domain. Additional implementation details and mathematical proof are provided in the supplementary material.

\subsection{Optimization in Equivariant Proxy Feature Space }
\paragraph{Instance-wise Alignment.} Here, we adopt a contrastive loss for local instance-wise optimization, which tends to produce a smooth content information transition from the source domain to the target domain within the equivariant proxy feature space. Specifically, given a batch of $N$ noise input $\lbrace \bm{z}_i \rbrace_{i=1}^N$, we could obtain the corresponding intermediate feature maps $\bm{I}^s_i$ and $\bm{I}^t_i$  in the source and target generators $\mathcal{G}_s(\cdot)$ and $\mathcal{G}_t(\cdot)$, respectively. The self-rotated instance-wise alignment loss is defined as follows:
\begin{equation}
    \mathcal{L}_{ins} \!= \!\sum_{i=1}^N \!-\!\log \!\frac{\exp(\text{sim}( \bm{I}^s_i, \exp(\hat{\bm{R}}-\hat{\bm{R}}^{\top})\bm{I}^t_i )/\tau )}{\sum_{j=1}^N \exp(\text{sim}( \bm{I}^s_i, \exp(\hat{\bm{R}}-\hat{\bm{R}}^{\top})\bm{I}^t_j )/\tau )},
\end{equation}
where $\text{sim}(\cdot,\cdot)$ denotes the cosine similarity, and $\tau$ is the temperature hyperparameter.

\paragraph{Distribution-level Matching.}
For the existing few-shot generative adaptation losses~\citep{RSSA}, the direct strong alignment of cross-domain pairwise relationships may result in an overemphasis of content information from the source domain. This over-saturation can introduce unwelcome information, thereby deteriorating the performance of the generation. To address this challenge, we shift our approach to embrace a global-scale optimal distribution matching strategy, grounded in optimal transport theory~\citep{ot1,ot2}. Specifically, we propose a distribution-level variational regularizer that penalizes the inter-domain distribution discrepancy via the intra-domain variations.
We first calculate the similarity graphs of intermediate features in the source and target domains $G_s = \lbrace \text{sim}(\bm{I}^s_i, \bm{I}^s_j) \rbrace_{i,j = 1}^N$ and $G_t = \lbrace \text{sim}(\bm{I}^t_k, \bm{I}^t_l) \rbrace_{k,l = 1}^N$, respectively. Then, we can indicate the inter-domain distribution discrepancy between the intermediate features in the source and target domains according to the discrepancy between intra-domain variations ${G}_s$ and $G_t$. In this procedure, we adopt the discrete optimal transport to measure the inter-domain discrepancy since it can effectively induce the intrinsic geometries of distributions. The corresponding Gromov Wasserstein distance between distributions $\mathcal{P}_s$ and $\mathcal{P}_t$ is formulated as follows:
\begin{equation}
\begin{aligned}
         \mathcal{W}(\mathcal{P}_s,\mathcal{P}_t)=& \sum_{i,j,k,l} \lvert  \text{sim}(\bm{I}^s_i, \bm{I}^s_j)
        -  \text{sim}( \exp(\hat{\bm{R}}-\hat{\bm{R}}^{\top})\bm{I}^t_k,\\
        &\exp(\hat{\bm{R}}-\hat{\bm{R}}^{\top})\bm{I}^t_l) \rvert^2\varLambda_{ik}\varLambda_{jl},
\end{aligned}
    \label{gwd}
\end{equation}
where $\lvert \cdot \rvert$ denotes the $\ell_1$ norm, $\varLambda_{ik}$ and $\varLambda_{jl}$ are the corresponding items of coupling matrix $\varLambda \in \mathbb{R}^{N \times N}$ that is constrained to satisfy $\varLambda\mathbbm{1}_N = \rho$ and $\varLambda^{\top}\mathbbm{1}_N = \varrho$, where $\mathbbm{1}_N$ denotes a $n$-dimensional all-one vector and $\rho, \varrho$ are weight vectors associated with $\bm^s_i, \bm{I}^t_k$. In this paper, we set $\rho_i = 1/N, \varrho_k = 1/N, i,k \in [1,2, c\dots, N]$.

\begin{table*}[t]
\centering
\renewcommand\arraystretch{0.9}
\setlength{\tabcolsep}{1.5mm}
\caption{The results of FID ($\downarrow$) of the few-shot image generation experiments on seven different target domains. The best results are indicated as \textbf{Bold}, and the second ones are indicated as \underline{Underline}.}
\begin{tabular}{llcccccccc}
\toprule
        & Backbone & B & S & MF & Skt & AC & AD & AW & Mean \\
\midrule
TGAN    & StyleGAN2  & 101.58 & 55.97      & 76.81    &    53.42      & 64.68    & 151.46   & 81.30 & 83.60    \\
TGAN+ADA & StyleGAN2  & 97.91  & 53.64      & 75.82    &   66.99       & 80.16    & 162.63   & 81.55  & 88.39   \\
FreezeD & StyleGAN2  & 96.25  & 46.95      & 73.33    &  46.54        & 63.60    & 157.98   & 77.18  & 80.26   \\
EWC     & StyleGAN2 & 79.93  & 49.41      & 62.67    & 64.55    & 74.61    & 158.78   & 92.83   & 83.25  \\
CDC      & StyleGAN2 & 69.13  & 41.45      & 65.45    & 47.62    & 176.21   & 170.95   & 135.13  & 100.85  \\
RSSA     & StyleGAN2 &  66.81      &   42.03         &   63.97       & 69.51    &    159.54      &  169.84        &     100.40     & 96.01 \\
SoLAD     & StyleGAN2 & 52.01  & 33.05      & 54.64    & 37.23    & 61.35    & 112.91   & 55.27   & 43.78  \\
AdAM     & StyleGAN2 & 48.83  & 28.03      & 51.34    & 42.64    & 58.07    & 100.91   & 36.87   & 52.38  \\
PIP     & StyleGAN2 & -  & 29.28      & -    & 37.40    & -    & -   & -   & -  \\
RICK     & StyleGAN2 & \underline{39.39}  & \underline{25.22}      & \underline{48.53}    & \underline{35.66}    & \textbf{53.27}    & \underline{98.71}    & \textbf{33.02}  & \underline{47.69}   \\
Ours     & StyleGAN2 & \textbf{37.16}  & \textbf{24.98}      & \textbf{46.03}    & \textbf{33.05}    & \underline{53.91}    & \textbf{97.22}    & \underline{34.31}  & \textbf{46.67}  \\
\midrule
DomainGallery      & SD1.4 & 58.86  & 43.10      & 60.38    & 44.86    & 77.15   & 123.54   & 65.32  & 78.22  \\

DDPM-PA      & DDPM & 48.92  & 34.75      & 55.39    & 39.68    & 69.22   & 58.27   & 60.24  & 52.35  \\
CRDI      & DDPM & 48.52  & 24.62      & 51.28    & 36.59    & 65.30   & 54.35   & 68.31  & 49.85  \\
DomainStudio   & DDPM & 33.26  & \textbf{21.92}      & -    & -    & -   & -   & -  & -  \\
Ours     & DDPM & \textbf{32.65}  & 22.98      & \textbf{31.44}    & \textbf{26.67}    & \underline{43.56}    & \textbf{77.52}    & \underline{28.15}  & \textbf{37.57}  \\

\midrule
$\text{RICK}^*$     & StyleGAN-ADA & 52.01  & 33.05      & 54.64    & 37.63    & 61.35    & 12.91   & 55.27   & 43.78  \\
WeditGAN     & StyleGAN-ADA & 48.83  & 28.03      & 51.34    & 35.41    & 58.07    & 100.91   & 36.87   & 52.38  \\
Ours     & StyleGAN-ADA & \textbf{32.74}  & \textbf{16.35}      & \textbf{40.02}    & \textbf{33.05}    & \underline{53.91}    & \textbf{97.22}    & \underline{34.31}  & \textbf{46.67}  \\
\bottomrule
\end{tabular}
\centering
\label{fid}
\end{table*}

Given that the solution for the distribution equalization, as described in Eq.(\ref{gwd}), poses a non-convex optimization problem, we employ the sliced Gromov Wasserstein distance for its resolution. Specifically, we project the learned metric space into various one-dimensional spaces using random directions. Through this approach, the sliced Gromov-Wasserstein distance can be effectively approximated by capturing sample observations from the distributions.
Formally, the sliced Gromov Wasserstein distance with $T$ projection vectors $\lbrace\pi_t\rbrace_{t=1}^T$ is easy to calculated as follows:
\begin{equation}
    \begin{aligned}
    \mathcal{L}_{dis} &= \frac{1}{T} \sum_{t=1}^T \sum_{i,j,k,l}| \text{sim}\left( \left<\bm{I}^s_i,\pi_t\right> , \left<\bm{I}^s_j,\pi_t\right> \right)\\
    &- \text{sim}(< \exp(\hat{\bm{R}}-\hat{\bm{R}}^{\top})\bm{I}^t_k,\pi_t>,\\
    &< \exp(\hat{\bm{R}}-\hat{\bm{R}}^{\top})\bm{I}^t_l,\pi_t>)|^2\varLambda_{ik}\varLambda_{jl}.
    \end{aligned}
\end{equation}

\paragraph{Overall.} The eventual loss function for optimization can be summarized as follows:
\begin{equation}
    \mathcal{L} = \mathcal{L}_{G_t} + \mathcal{L}_{D} + \lambda_1\mathcal{L}_{ins} + \lambda_2\mathcal{L}_{dis},
\end{equation}
where $\lambda_1, \lambda_2$ are the hyperparameters.

\begin{figure*}[t]
    \centering
\includegraphics[width = 0.88\textwidth]{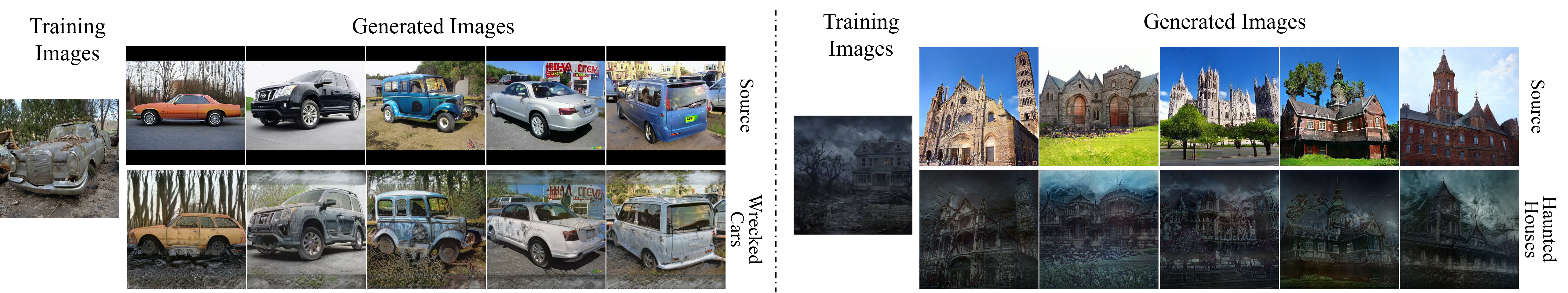}
\caption{Generative adaptation results of our method on \texttt{Cars} $\rightarrow$ \texttt{Wrecked cars} and \texttt{Church} $\rightarrow$ \texttt{Haunted house} (10-shot). Best zoomed in and viewed in color.}
    \label{fig:vis2}
\end{figure*}

\begin{figure*}[t]
	\centering
	\includegraphics[scale=0.39]{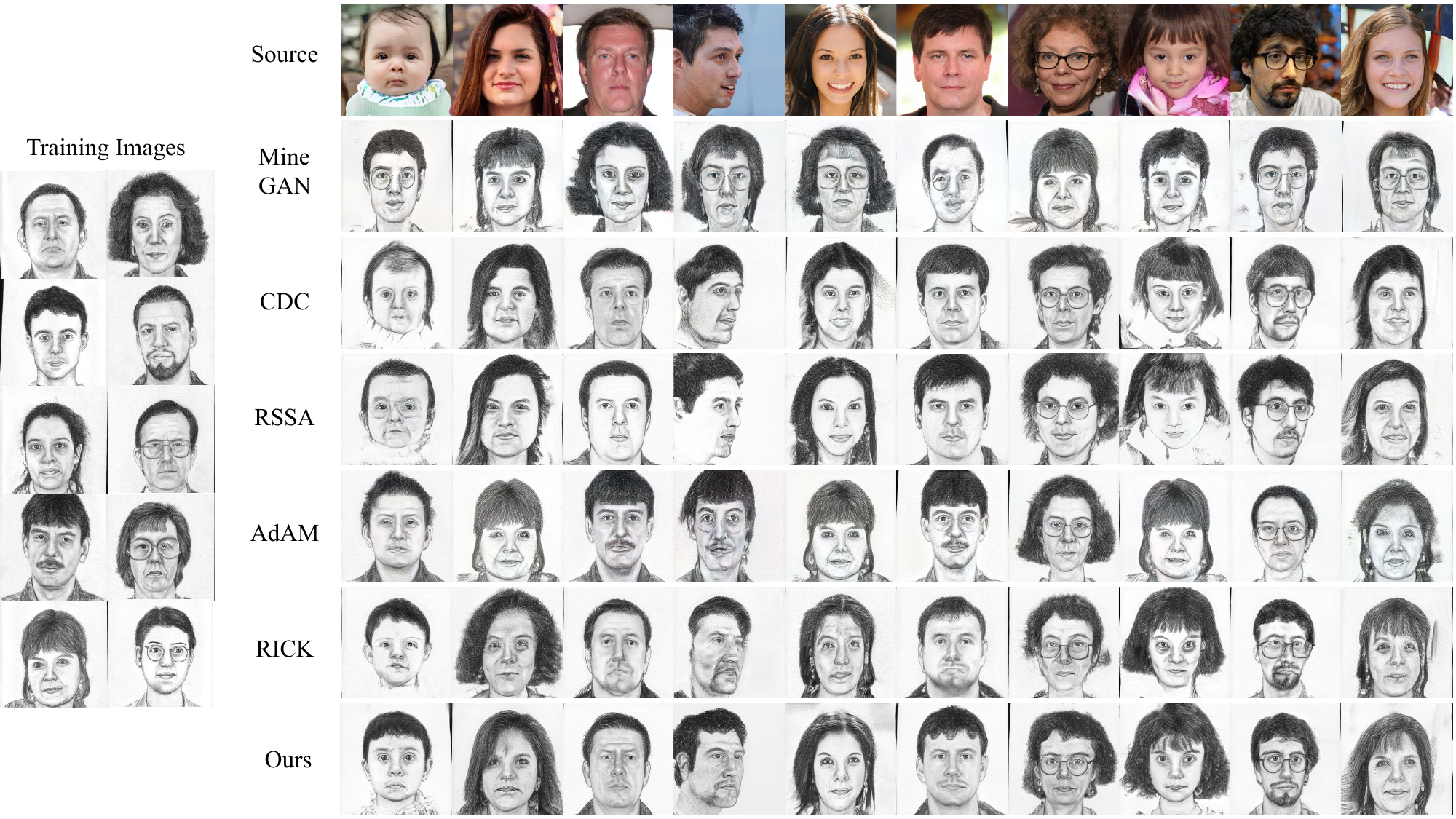}
	\\		
\vspace{3pt}
 \includegraphics[scale=0.39]{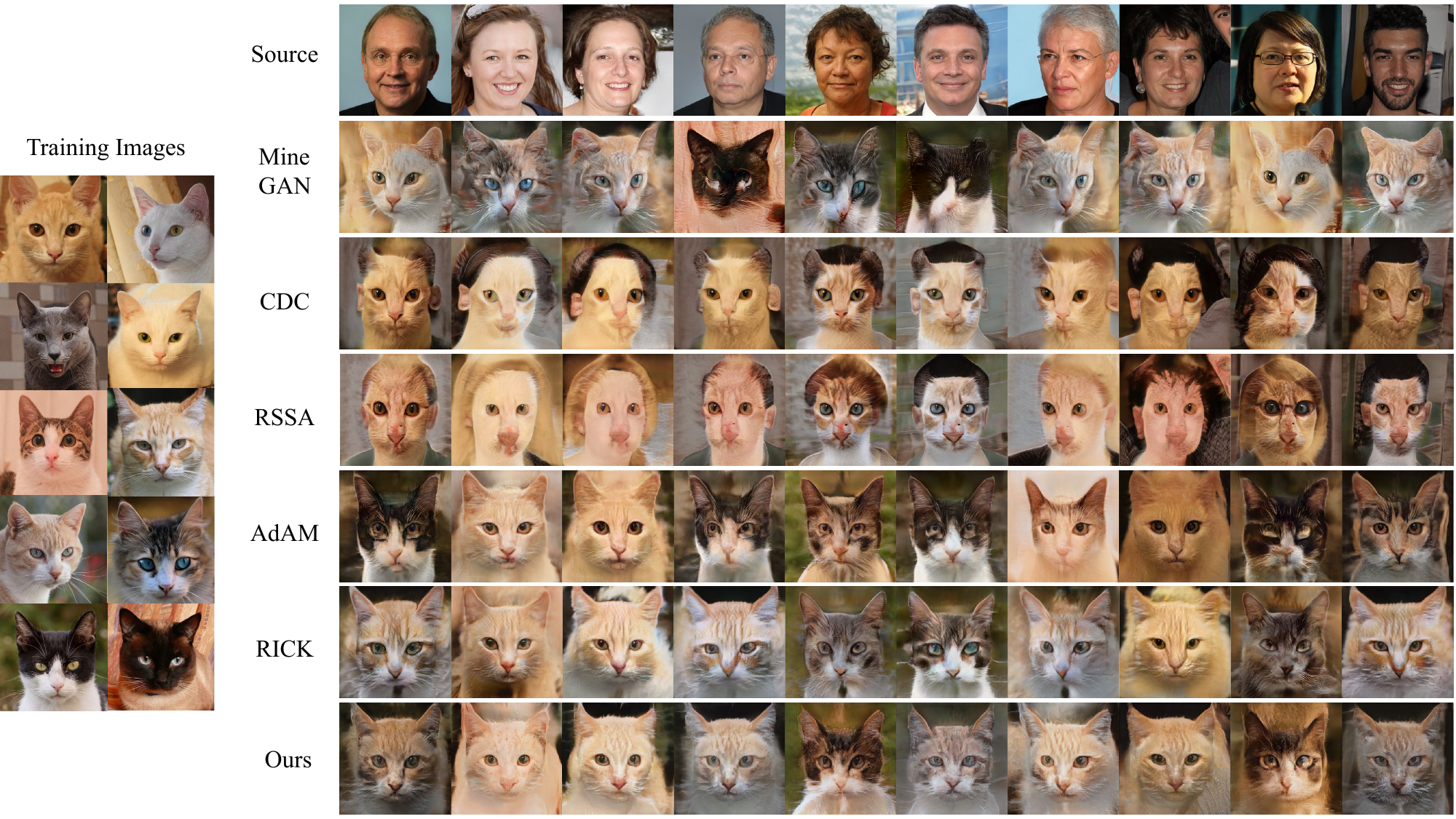} 
	\caption{ Visualized comparison results with different methods on \texttt{FFHQ} $\rightarrow$ \texttt{Sketches} and  \texttt{FFHQ} $\rightarrow$ \texttt{AFHQ-Cat} adaptations. The target training data is under the 10-shot setting. Synthesized samples in each column are generated with the same random input $\mathbf{z}$. Best zoomed in and viewed in color. Best zoomed in and viewed in color.}
	\label{fig:vis} 
\end{figure*}

\begin{figure*}[t]
    \centering
\includegraphics[width = 0.88\textwidth]{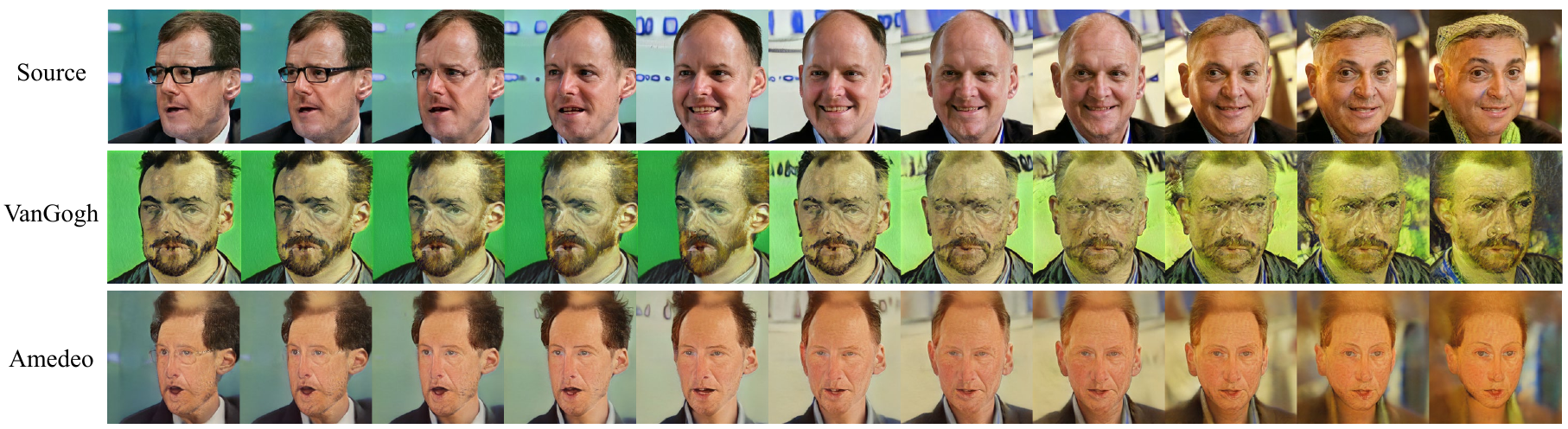}
    \caption{Latent space interpolation results on \texttt{FFHQ$\rightarrow$VanGoGh} and \texttt{FFHQ$\rightarrow$Amedeo}}
    \label{fig:chazhi}
\end{figure*}

\section{Experiments}

In this section, we evaluate our proposed SeAM and analyze its essential characteristics. 

\subsection{Experimental Settings}

\textbf{Experimental Implementation.} For a fair comparison, we follow the standard experimental protocol~\citep{CDC,AdAM} as previous works and explore different $source \rightarrow target$ adaptation settings to analyze the effectiveness of our method. Model adaptations are done in a 10-shot setting. In all experiments, StyleGAN-V2~\citep{stylegan_v2} is employed as the GAN architecture for pre-training and fine-tuning. We train our models on an NVIDIA A6000. We operate on 256 × 256 resolution images for both the source and target domains. 
We train the generator and discriminator by Adam optimizer~\citep{ADAMopt} with the same hyperparameters (learning rate, $\beta_1$, and $\beta_2$ are set as 0.002, 0, and 0.99, respectively) as previous works. 
We set $\lambda_1$ and $\lambda_2$ to 0.6 and 0.4 in all the experiments.
We set the size of the mini-batch as 8 and trained for about 1000 iterations. 
To comprehensively evaluate the effectiveness of the proposed model, we compare it with various state-of-the-art methods based on different backbones. Specifically, we include methods built upon Diffusion Models (DMs): DomainGallery~\cite{duan2024domaingallery}, DDPM-PA~\cite{zhu2022few}, CRDI~\cite{cao2024few}, and DomainStudio~\cite{zhu2025domainstudio}; a method based on StyleGAN-ADA~\cite{karras2020training}: WeditGAN~\cite{duan2024weditgan}; as well as methods based on the StyleGAN2~\cite{karras2020analyzing} backbone: TGAN~\citep{TGAN}, TGAN+ADA~\citep{TGANada}, FreezeD~\citep{FreezeD}, EWC~\citep{EWC}, MineGAN~\citep{minegan}, CDC~\citep{CDC}, RSSA~\citep{RSSA}, SoLAD~\cite{mondal2024solad}, AdAM~\citep{AdAM}, PIP~\cite{li2024peer}, and RICK~\citep{RICK}.

\textbf{Datasets.} Following previous works, we use the GAN pre-trained on a large-scale image dataset FFHQ~\citep{karras2019style}. For few-shot adaptations, we select several target domains that have different proximity to the source dataset: 1) Semantic-related domains: Babies(B)~\citep{CDC}, Sunglasses(S)~\citep{CDC}, Sketches(Skt)~\citep{wang2008face}, and MetFaces(MF)~\citep{TGANada}; 2) Distant domains: AFHQ-Cat(AC)~\citep{choi2020stargan}, AFHQ-Dog(AD)~\citep{choi2020stargan}, and AFHQ-Wild(AW)~\citep{choi2020stargan}.

To augment the evaluation of the efficacy of the sophisticated SeAM methodology, we incorporate two supplementary experimental trials for alternate source domains. We select LSUN Churches~\citep{yu2015lsun} and LSUN-Stanford car~\citep{lsuncar} datasets as source domain datasets. Subsequently, we correspondingly adapt them to the haunted house~\citep{CDC} and wrecked cars~\citep{CDC} datasets.

\textbf{Evaluation Metrics.} For a more robust demonstration of the effectiveness of our proposed methodology, we utilize three separate evaluation methods to measure not only the quality but also the diversity of the images generated by our technique in conjunction with those created by the comparative baselines. We employ the Fr\'{e}chet Inception Distance (FID)~\citep{fid}, a widely accepted metric, to quantify the divergence between the fitted Gaussian distribution of authentic and generated samples. Additionally, we utilize the intra-cluster variant of the Learned Perceptual Image Patch Similarity (intra-LPIPS) to ascertain the variance within the collection of images generated by our method~\citep{CDC}. As a final point, we visually display the generated images to provide a more instinctive comparison.

\subsection{Evaluation Results}

\textbf{Qualitative Result.} The generated samples on \texttt{FFHQ} $\rightarrow$ \texttt{Sketches} and \texttt{FFHQ} $\rightarrow$ \texttt{AFHQ-Cats}, employing varying few-shot generative adaption methods are shown in Figure~\ref{fig:vis}. In the adaptation \texttt{FFHQ} $\rightarrow$ \texttt{Sketches}, MineGAN demonstrates a significant overfitting to the samples derived from the target domain. When juxtaposed with MineGAN, AdAM does not yield any enhancement in the outcomes. In the adaptation process \texttt{FFHQ} $\rightarrow$ \texttt{AFHQ-Cats}, both RSSA and CDC exhibit underfitting towards the target source, implying that the generated feline images via these methods still maintain certain human traits. As evidenced by the generated images, our proposed SeAM method delivers superior generation performance due to its ability to adeptly inherit the characteristics while effectively encapsulating those from the target domains. This underscores the efficacy of our unified knowledge embedding strategy.

To further substantiate the universal applicability of our methodology, we executed experiments in a 10-shot setting across three distinct scenarios: \texttt{Cars} $\rightarrow$ \texttt{Wrecked cars}  \texttt{Church} $\rightarrow$ \texttt{Haunted house}. The empirical in Figure~\ref{fig:vis2} outcomes indicate that our method exhibits exceptional performance irrespective of the pre-training models utilized.

\begin{table*}[t]
\centering

\begin{minipage}{0.48\textwidth}
\centering
\renewcommand\arraystretch{0.86}
\caption{The results of intra-LPIPS ($\uparrow$) on three target domains.}
\begin{tabular}{lccc}
\toprule
         & B & AC  & Skt\\
\midrule
TGAN     & 0.517  & 0.490  & 0.386  \\
TGAN+ADA & 0.511  & 0.513  & 0.344  \\
FreezeD  & 0.518  & 0.492  & 0.351  \\
EWC      & 0.521  & 0.587  & 0.423  \\
CDC      & 0.578  & \textbf{0.629}     & 0.418 \\
RSSA     & 0.582  & \underline{0.612}  & 0.478    \\
SoLAD    & 0.587  & 0.601  & 0.483 \\
AdAM     & 0.590  & 0.557              & 0.482   \\
RICK     & \underline{0.608}  & 0.569  & 0.493    \\
DomainStudio     & 0.599  & -  & \underline{0.495}    \\
Ours     & \textbf{0.613}  & \underline{0.612}  & \textbf{0.511}  \\
\bottomrule
\end{tabular}
\label{lpips}
\end{minipage}
\hfill
\begin{minipage}{0.48\textwidth}
\centering
\renewcommand\arraystretch{1.05}
\setlength{\tabcolsep}{1mm}
\caption{The ablation study of different losses and the effect of rotation for EFR. $R^*$ indicates whether the rotation strategy is applied. Results are evaluated using FID ($\downarrow$). }
\begin{tabular}{ccc|ccc}
\toprule
$R^*$ &$\mathcal{L}_{ins}$ & $\mathcal{L}_{dis}$ & B & S & MF \\
\midrule
       & \checkmark          &                      & 54.26  & 41.77      & 65.37    \\
\checkmark & \checkmark       &                      & 43.91  & 37.41      & 60.32    \\
       &                      & \checkmark           & 51.35  & 40.38      & 64.18    \\
\checkmark &                  & \checkmark           & 42.88  & 34.69      & 62.14    \\
       & \checkmark          & \checkmark           & 46.33  & 29.05      & 54.64    \\
\checkmark & \checkmark       & \checkmark           & \textbf{37.16} & \textbf{24.98} & \textbf{46.03} \\
\bottomrule
\end{tabular}
\label{tab:ab_loss}
\end{minipage}

\end{table*}

To provide a comprehensive demonstration of the efficacy of our Equivariant Self-Rotated Proxy Feature Space Projection strategy, we also undertake a comparison of latent space interpolation. We subtly adjust the metric by selecting 10 subintervals between any two latent vectors, thereby generating corresponding images by source model. Concurrently, we generate images using the target model with the same latent vectors. 
As shown in Figure~\ref{fig:chazhi}, it becomes evident that the data distribution within the target domain aligns with the source domain. For instance, the rotation of a person's head in the source domain corresponds to the angle in the target domain. 
This demonstrates that the proposed proxy space enables a smooth and stable alignment of spatial distributions.

\begin{figure*}[t]
    \centering
    \begin{minipage}[b]{0.67\textwidth}
        \centering
        \includegraphics[width=\textwidth]{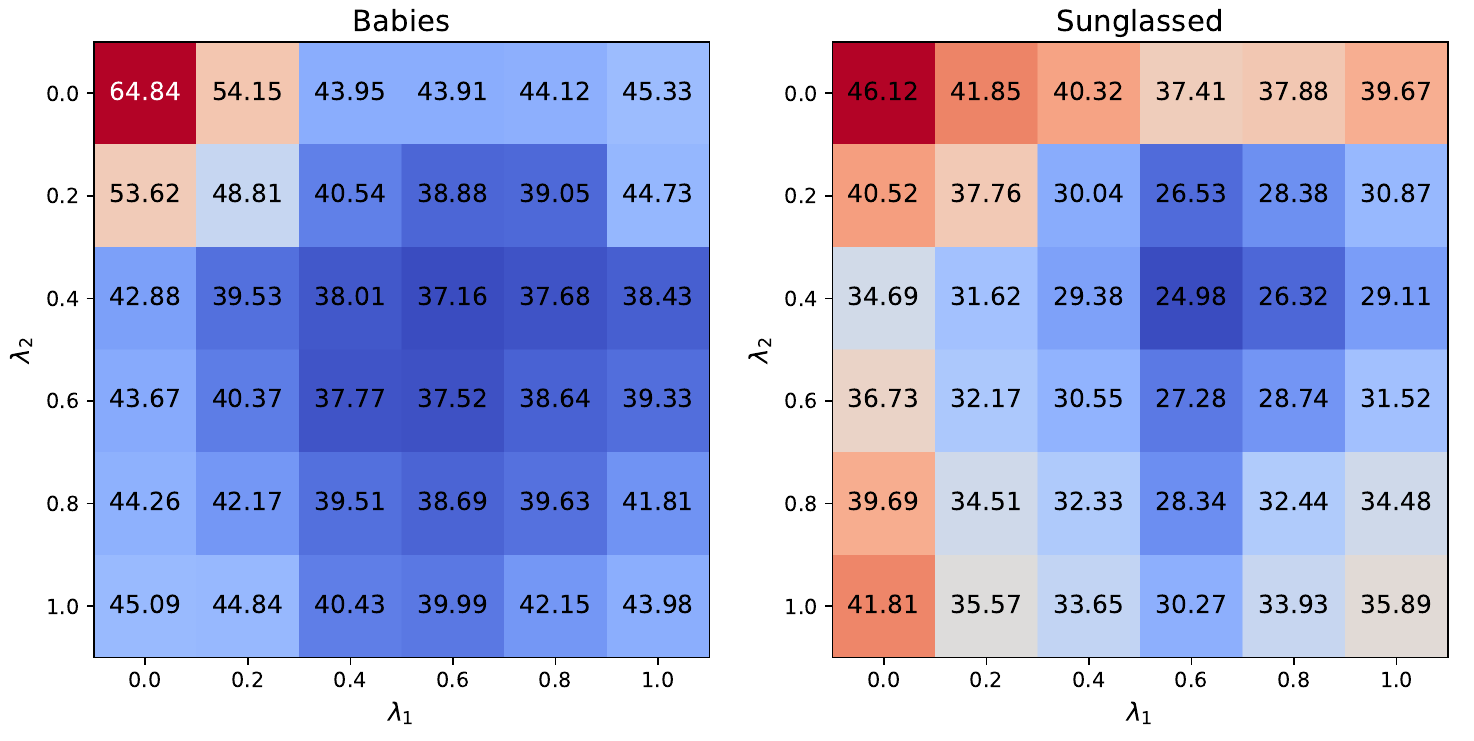}
        \caption{The FID results with regard to $\lambda_1$ and $\lambda_2$ on different datasets.}
        \label{fig:lambda1}
    \end{minipage}
    \hfill
    \begin{minipage}[b]{0.26\textwidth}
        \centering
        \includegraphics[width=\textwidth]{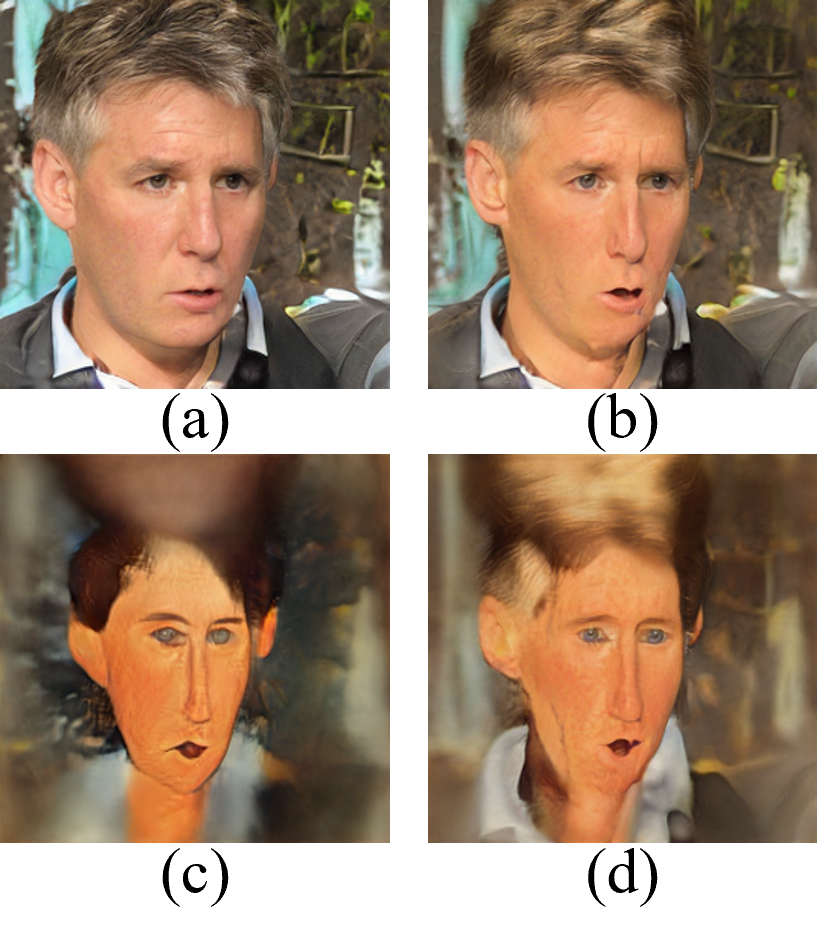}
        \caption{The generative visualization of rotation.}
        \label{fig:lambda2}
    \end{minipage}
 \vspace{-15pt}
\end{figure*}

To more comprehensively validate the effectiveness of our self-adaptive rotation strategy, we visualize the generative outcomes of the rotated features, as illustrated in Figure~\ref{fig:lambda2}. Specifically, subfigure (a) depicts the image generated from the source feature using the source generator, $\mathcal{G}_s(\bm{I}^s)$; (b) shows the output $\mathcal{G}_s(\bm{R}^* \bm{I}^t)$, where the target feature is first rotated; (c) presents the generated image $\mathcal{G}_t(\bm{I}^t)$ from the target generator; and (d) illustrates $\mathcal{G}_t((\bm{R}^*)^{\top} \bm{I}^s)$, where the source feature is inversely rotated.
The visual comparison indicates that the rotation matrix primarily transfers domain-specific characteristics without introducing additional semantic content. This supports our claim that the self-rotated proxy feature space effectively bridges the domain gap while preserving the intrinsic content of the original representations.

\textbf{Quantitative Result.} In order to further delineate the caliber and heterogeneity of the synthesized images, the Fr$\Acute{e}$chet Inception Distance (FID)~\citep{fid} and intra-cluster version of the Learned Perceptual Image Patch Similarity (intra-LPIPS)~\citep{CDC} scores are utilized as quantitative measures. The comprehensive numerical findings pertaining to several benchmark datasets are documented in Table~\ref{fid} and Table~\ref{lpips}. The target datasets comprise a significant volume of images, for instance, 5000 instances in the AFHQ-Cat dataset. It is feasible to employ the entire target dataset for evaluative purposes by using our refined generator to generate an equivalent quantity of images arbitrarily. Among the entirety of these outcomes, our methodology exhibits superior performance on both metrics, thereby evidencing the efficacy of our proposed technique.

\subsection{Ablation Study}



\textbf{The Impact of Diverse Compositional Losses and Rotation.}
To systematically evaluate the influence of different loss components, we conduct an ablation study considering various combinations of $\mathcal{L}_{ins}$ and $\mathcal{L}_{dis}$. The resulting FID scores are summarized in Table~\ref{tab:ab_loss}. It is evident that both losses contribute positively to the performance of the generation model.
To further assess the effectiveness of the rotation mechanism, additional ablation experiments are carried out. As shown in Table~\ref{tab:ab_loss}, incorporating rotation consistently improves performance across all settings, demonstrating its beneficial impact on few-shot image generation.

\textbf{The Impact of different weights for Losses.} To meticulously analyze the influence of assorted $\lambda_1$ and $\lambda_2$ variables, we conduct ablation studies on the hyperparameters $\lambda_1$ and $\lambda_2$ using a grid search approach over the Babies and Sunglasses datasets. As shown in the corresponding Figure~\ref{fig:lambda1}, the performance drops significantly when both $\lambda_1$ and $\lambda_2$ are set to zero, indicating the importance of the proposed losses. 
It is evident that the quantitative values of the hyperparameters exert minimal effect on the outcomes, implying that the $\mathcal{L}_{ins}$ and $\mathcal{L}_{dis}$ sensitivity are not pronounced. 
Based on the results, we select $\lambda_1 = 0.6$ and $\lambda_2 = 0.4$ as the final hyperparameters configuration.

\section{Conclusion} 

This paper presents Equivariant Feature Rotation (EFR), a novel and effective strategy for few-shot generative adaptation. By introducing a self-rotated proxy feature space through learnable Lie Group-based rotations, EFR addresses the fundamental challenge of misaligned distribution patterns between source and target domains. This approach not only preserves the structural integrity of intra-domain representations but also facilitates robust cross-domain alignment, enabling more stable and informative generation. Extensive empirical evaluations across multiple benchmarks confirm that EFR significantly mitigates overfitting and content degradation, outperforming existing methods in terms of both content fidelity and adaptation efficiency. These findings underscore the potential of equivariant transformations in advancing the capabilities of few-shot generative modeling.

{
    \small
    \bibliographystyle{ieeenat_fullname}
    \bibliography{main}

@String(PAMI = {IEEE Trans. Pattern Anal. Mach. Intell.})

@String(CVPR= {IEEE Conf. Comput. Vis. Pattern Recog.})

@String(ICCV= {Int. Conf. Comput. Vis.})

@String(BMVC= {Brit. Mach. Vis. Conf.})

@String(TIP  = {IEEE Trans. Image Process.})

@String(ICLR = {Int. Conf. Learn. Represent.})

@String(AAAI = {AAAI})

@String(PAMI  = {IEEE TPAMI})

@String(CVPR  = {CVPR})

@String(ICCV  = {ICCV})

@String(BMVC  =	{BMVC})

@String(TIP   = {IEEE TIP})

@String(ICLR  = {ICLR})

@article{zhang2019progressive,
  title={Progressive augmentation of gans},
  author={Zhang, Dan and Khoreva, Anna},
  journal={NeurIPS},
  year={2019}
}

@inproceedings{zhao2021improved,
  title={Improved consistency regularization for gans},
  author={Zhao, Zhengli and Singh, Sameer and Lee, Honglak and Zhang, Zizhao and Odena, Augustus and Zhang, Han},
  booktitle={AAAI},
  year={2021}
}

@article{zhao2020image,
  title={Image augmentations for gan training},
  author={Zhao, Zhengli and Zhang, Zizhao and Chen, Ting and Singh, Sameer and Zhang, Han},
  journal={arXiv},
  year={2020}
}

@article{ot1,
  title={Computational optimal transport: With applications to data science},
  author={Peyr{\'e}, Gabriel and Cuturi, Marco and others},
  journal={Foundations and Trends{\textregistered} in Machine Learning},
  volume={11},
  number={5-6},
  pages={355--607},
  year={2019},
  publisher={Now Publishers, Inc.}
}

@article{ot2,
  title={Optimal transport on discrete domains},
  author={Solomon, Justin},
  journal={AMS Short Course on Discrete Differential Geometry},
  year={2018},
  publisher={Notices of the AMS}
}

@inproceedings{lezcano2019cheap,
  title={Cheap orthogonal constraints in neural networks: A simple parametrization of the orthogonal and unitary group},
  author={Lezcano-Casado, Mario and Mart{\i}nez-Rubio, David},
  booktitle={ICML},
  pages={3794--3803},
  year={2019},
  organization={PMLR}
}

@article{quan2022image,
  title={Image inpainting with local and global refinement},
  author={Quan, Weize and Zhang, Ruisong and Zhang, Yong and Li, Zhifeng and Wang, Jue and Yan, Dong-Ming},
  journal={IEEE TIP},
  year={2022}
}

@article{cao2021unifacegan,
  title={UniFaceGAN: a unified framework for temporally consistent facial video editing},
  author={Cao, Meng and Huang, Haozhi and Wang, Hao and Wang, Xuan and Shen, Li and Wang, Sheng and Bao, Linchao and Li, Zhifeng and Luo, Jiebo},
  journal={IEEE TIP},
  year={2021}
}

@article{zhang2022survey,
  title={A survey on multimodal-guided visual content synthesis},
  author={Zhang, Ziqi and Li, Zeyu and Wei, Kun and Pan, Siduo and Deng, Cheng},
  journal={Neurocomputing},
  year={2022},
  publisher={Elsevier}
}

@article{brock2018large,
  title={Large scale GAN training for high fidelity natural image synthesis},
  author={Brock, Andrew and Donahue, Jeff and Simonyan, Karen},
  journal={ICLR},
  year={2018}
}

@inproceedings{karras2019style,
  title={A style-based generator architecture for generative adversarial networks},
  author={Karras, Tero and Laine, Samuli and Aila, Timo},
  booktitle={CVPR},
  year={2019}
}

@inproceedings{stylegan_v2,
  title={Analyzing and improving the image quality of stylegan},
  author={Karras, Tero and Laine, Samuli and Aittala, Miika and Hellsten, Janne and Lehtinen, Jaakko and Aila, Timo},
  booktitle={CVPR},
  year={2020}
}

@inproceedings{yeh2017semantic,
  title={Semantic image inpainting with deep generative models},
  author={Yeh, Raymond A and Chen, Chen and Yian Lim, Teck and Schwing, Alexander G and Hasegawa-Johnson, Mark and Do, Minh N},
  booktitle={CVPR},
  year={2017}
}

@inproceedings{yu2018generative,
  title={Generative image inpainting with contextual attention},
  author={Yu, Jiahui and Lin, Zhe and Yang, Jimei and Shen, Xiaohui and Lu, Xin and Huang, Thomas S},
  booktitle={CVPR},
  year={2018}
}

@inproceedings{zhu2017unpaired,
  title={Unpaired image-to-image translation using cycle-consistent adversarial networks},
  author={Zhu, Jun-Yan and Park, Taesung and Isola, Phillip and Efros, Alexei A},
  booktitle={ICCV},
  year={2017}
}

@article{yu2015lsun,
  title={Lsun: Construction of a large-scale image dataset using deep learning with humans in the loop},
  author={Yu, Fisher and Seff, Ari and Zhang, Yinda and Song, Shuran and Funkhouser, Thomas and Xiao, Jianxiong},
  journal={arXiv},
  year={2015}
}

@article{huang2022masked,
  title={Masked generative adversarial networks are data-efficient generation learners},
  author={Huang, Jiaxing and Cui, Kaiwen and Guan, Dayan and Xiao, Aoran and Zhan, Fangneng and Lu, Shijian and Liao, Shengcai and Xing, Eric},
  journal={NeurIPS},
  year={2022}
}

@article{tran2021data,
  title={On data augmentation for gan training},
  author={Tran, Ngoc-Trung and Tran, Viet-Hung and Nguyen, Ngoc-Bao and Nguyen, Trung-Kien and Cheung, Ngai-Man},
  journal={IEEE TIP},
  year={2021}
}

@inproceedings{tseng2021regularizing,
  title={Regularizing generative adversarial networks under limited data},
  author={Tseng, Hung-Yu and Jiang, Lu and Liu, Ce and Yang, Ming-Hsuan and Yang, Weilong},
  booktitle={CVPR},
  year={2021}
}

@article{yang2021data,
  title={Data-efficient instance generation from instance discrimination},
  author={Yang, Ceyuan and Shen, Yujun and Xu, Yinghao and Zhou, Bolei},
  journal={NeurIPS},
  year={2021}
}

@inproceedings{zhao2020leveraging,
  title={On leveraging pretrained GANs for generation with limited data},
  author={Zhao, Miaoyun and Cong, Yulai and Carin, Lawrence},
  booktitle={ICML},
  year={2020}
}

@inproceedings{li2023task,
  title={Task-specific fine-tuning via variational information bottleneck for weakly-supervised pathology whole slide image classification},
  author={Li, Honglin and Zhu, Chenglu and Zhang, Yunlong and Sun, Yuxuan and Shui, Zhongyi and Kuang, Wenwei and Zheng, Sunyi and Yang, Lin},
  booktitle={CVPR},
  year={2023}
}

@article{pan2009survey,
  title={A survey on transfer learning},
  author={Pan, Sinno Jialin and Yang, Qiang},
  journal={IEEE TKDE},
  year={2009}
}

@article{zhuang2020comprehensive,
  title={A comprehensive survey on transfer learning},
  author={Zhuang, Fuzhen and Qi, Zhiyuan and Duan, Keyu and Xi, Dongbo and Zhu, Yongchun and Zhu, Hengshu and Xiong, Hui and He, Qing},
  journal={Proceedings of the IEEE},
  year={2020},
  publisher={IEEE}
}

@inproceedings{TGAN,
  title={Convolutional neural network pruning with structural redundancy reduction},
  author={Wang, Zi and Li, Chengcheng and Wang, Xiangyang},
  booktitle={CVPR},
  year={2021}
}

@article{TGANada,
  title={Training generative adversarial networks with limited data},
  author={Karras, Tero and Aittala, Miika and Hellsten, Janne and Laine, Samuli and Lehtinen, Jaakko and Aila, Timo},
  journal={NeurIPS},
  year={2020}
}

@inproceedings{CDC,
  title={Few-shot image generation via cross-domain correspondence},
  author={Ojha, Utkarsh and Li, Yijun and Lu, Jingwan and Efros, Alexei A and Lee, Yong Jae and Shechtman, Eli and Zhang, Richard},
  booktitle={CVPR},
  year={2021}
}

@article{AdAM,
  title={Few-shot image generation via adaptation-aware kernel modulation},
  author={Zhao, Yunqing and Chandrasegaran, Keshigeyan and Abdollahzadeh, Milad and Cheung, Ngai-Man Man},
  journal={NeurIPS},
  year={2022}
}

@article{EWC,
  title={Few-shot image generation with elastic weight consolidation},
  author={Li, Yijun and Zhang, Richard and Lu, Jingwan and Shechtman, Eli},
  journal={arXiv},
  year={2020}
}

@inproceedings{DWSC,
  title={Dynamic Weighted Semantic Correspondence for Few-Shot Image Generative Adaptation},
  author={Hou, Xingzhong and Liu, Boxiao and Zhang, Shuai and Shi, Lulin and Jiang, Zite and You, Haihang},
  booktitle={ACM MM},
  year={2022}
}

@inproceedings{RSSA,
  title={Few shot generative model adaption via relaxed spatial structural alignment},
  author={Xiao, Jiayu and Li, Liang and Wang, Chaofei and Zha, Zheng-Jun and Huang, Qingming},
  booktitle={CVPR},
  year={2022}
}

@inproceedings{DCL,
  title={A closer look at few-shot image generation},
  author={Zhao, Yunqing and Ding, Henghui and Huang, Houjing and Cheung, Ngai-Man},
  booktitle={CVPR},
  year={2022}
}

@inproceedings{RICK,
    title={Exploring incompatible knowledge transfer in few-shot image generation},
    author={Zhao, Yunqing and Du, Chao and Abdollahzadeh, Milad and Pang, Tianyu and Lin, Min and Yan, Shuicheng and Cheung, Ngai-Man},
    booktitle={CVPR},
    year={2023}
}

@article{FreezeD,
  title={Freeze the discriminator: a simple baseline for fine-tuning gans},
  author={Mo, Sangwoo and Cho, Minsu and Shin, Jinwoo},
  journal={arXiv},
  year={2020}
}

@inproceedings{minegan,
  title={Minegan: effective knowledge transfer from gans to target domains with few images},
  author={Wang, Yaxing and Gonzalez-Garcia, Abel and Berga, David and Herranz, Luis and Khan, Fahad Shahbaz and Weijer, Joost van de},
  booktitle={CVPR},
  year={2020}
}

@inproceedings{choi2020stargan,
  title={Stargan v2: Diverse image synthesis for multiple domains},
  author={Choi, Yunjey and Uh, Youngjung and Yoo, Jaejun and Ha, Jung-Woo},
  booktitle={CVPR},
  pages={8188--8197},
  year={2020}
}

@article{bounareli2022finding,
  title={Finding Directions in GAN’s Latent Space for Neural Face Reenactment},
  author={Bounareli, Stella and Argyriou, Vasileios and Tzimiropoulos, Georgios},
  journal={BMVC},
  year={2022}
}

@inproceedings{collins2020editing,
  title={Editing in style: Uncovering the local semantics of gans},
  author={Collins, Edo and Bala, Raja and Price, Bob and Susstrunk, Sabine},
  booktitle={CVPR},
  year={2020}
}

@article{wang2008face,
  title={Face photo-sketch synthesis and recognition},
  author={Wang, Xiaogang and Tang, Xiaoou},
  journal={IEEE PAMI},
  year={2008}
}

@article{fid,
  title={Gans trained by a two time-scale update rule converge to a local nash equilibrium},
  author={Heusel, Martin and Ramsauer, Hubert and Unterthiner, Thomas and Nessler, Bernhard and Hochreiter, Sepp},
  journal={NeurIPS},
  year={2017}
}

@article{SGD,
  title={A stochastic approximation method},
  author={Robbins, Herbert and Monro, Sutton},
  journal={The annals of mathematical statistics},
  pages={400--407},
  year={1951},
  publisher={JSTOR}
}

@article{ADAMopt,
  title={Adam: A method for stochastic optimization},
  author={Kingma, Diederik P and Ba, Jimmy},
  journal={arXiv preprint arXiv:1412.6980},
  year={2014}
}

@article{lsuncar,
  title={LSUN-Stanford car dataset: enhancing large-scale car image datasets using deep learning for usage in GAN training},
  author={Kramberger, Tin and Poto{\v{c}}nik, Bo{\v{z}}idar},
  journal={Applied Sciences},
  volume={10},
  number={14},
  pages={4913},
  year={2020},
  publisher={MDPI}
}

@article{li2024peer,
  title={Peer is Your Pillar: A Data-unbalanced Conditional GANs for Few-shot Image Generation},
  author={Li, Ziqiang and Wang, Chaoyue and Rui, Xue and Xue, Chao and Leng, Jiaxu and Fu, Zhangjie and Li, Bin},
  journal={IEEE Transactions on Circuits and Systems for Video Technology},
  year={2024},
  publisher={IEEE}
}

@article{mondal2024solad,
  title={SoLAD: Sampling over Latent Adapter for Few Shot Generation},
  author={Mondal, Arnab Kumar and Tiwary, Piyush and Singla, Parag and Prathosh, AP},
  journal={IEEE Signal Processing Letters},
  year={2024},
  publisher={IEEE}
}

@inproceedings{duan2024weditgan,
  title={Weditgan: Few-shot image generation via latent space relocation},
  author={Duan, Yuxuan and Niu, Li and Hong, Yan and Zhang, Liqing},
  booktitle={Proceedings of the AAAI Conference on Artificial Intelligence},
  volume={38},
  number={2},
  pages={1653--1661},
  year={2024}
}

@article{zhu2022few,
  title={Few-shot image generation with diffusion models},
  author={Zhu, Jingyuan and Ma, Huimin and Chen, Jiansheng and Yuan, Jian},
  journal={arXiv preprint arXiv:2211.03264},
  year={2022}
}

@inproceedings{cao2024few,
  title={Few-shot image generation by conditional relaxing diffusion inversion},
  author={Cao, Yu and Gong, Shaogang},
  booktitle={European Conference on Computer Vision},
  pages={20--37},
  year={2024},
  organization={Springer}
}

@article{duan2024domaingallery,
  title={DomainGallery: Few-shot Domain-driven Image Generation by Attribute-centric Finetuning},
  author={Duan, Yuxuan and Hong, Yan and Zhang, Bo and Zhu, Huijia and Wang, Weiqiang and Zhang, Jianfu and Niu, Li and Zhang, Liqing and others},
  journal={Advances in Neural Information Processing Systems},
  volume={37},
  pages={537--561},
  year={2024}
}

@article{karras2020training,
  title={Training generative adversarial networks with limited data},
  author={Karras, Tero and Aittala, Miika and Hellsten, Janne and Laine, Samuli and Lehtinen, Jaakko and Aila, Timo},
  journal={Advances in neural information processing systems},
  volume={33},
  pages={12104--12114},
  year={2020}
}

@inproceedings{karras2020analyzing,
  title={Analyzing and improving the image quality of stylegan},
  author={Karras, Tero and Laine, Samuli and Aittala, Miika and Hellsten, Janne and Lehtinen, Jaakko and Aila, Timo},
  booktitle={Proceedings of the IEEE/CVF conference on computer vision and pattern recognition},
  pages={8110--8119},
  year={2020}
}

@article{zhu2025domainstudio,
  title={DomainStudio: Fine-Tuning Diffusion Models for Domain-Driven Image Generation Using Limited Data},
  author={Zhu, Jingyuan and Ma, Huimin and Chen, Jiansheng and Yuan, Jian},
  journal={International Journal of Computer Vision},
  volume={133},
  number={10},
  pages={7012--7036},
  year={2025},
  publisher={Springer}
}
}


\end{document}